\documentclass[10pt,twocolumn,letterpaper]{article}
\usepackage{iccv}
\usepackage{times}
\usepackage[pdftex]{graphicx}
\usepackage{color}
\usepackage{amsmath,amssymb}
\usepackage{algorithm,algpseudocode}
\usepackage{subfig}
\usepackage{units}
\pdfoutput=1
\graphicspath{{./figures/}}
\DeclareGraphicsExtensions{.pdf,.jpeg,.png,.jpg}

\usepackage[breaklinks=true,bookmarks=false,pagebackref=true,colorlinks=true]{hyperref}

\usepackage{xspace}
\def\sota{state-of-the-art}

\def\matlab{MATLAB}

\newcommand{\refsec}[1]{Section~\ref{#1}}
\newcommand{\reffig}[1]{Figure~\ref{#1}}
\newcommand{\refeq}[1]{Equation~\eqref{#1}}
\newcommand{\refalg}[1]{Algorithm~\ref{#1}}
\newcommand{\reftab}[1]{Table~\ref{#1}}

\renewcommand{\vec}[1]{\mathbf{#1}}
\newcommand{\set}[1]{\mathcal{#1}}

\newcommand{\R}{\mathbb{R}}

\DeclareMathOperator*{\argmin}{arg\,min}

\newcommand{\p}[1]{\vec{p}_{#1}} 	 
\newcommand{\f}[1]{\vec{f}_{#1}} 	 
\newcommand{\g}[1]{\vec{g}_{#1}} 	 
\newcommand{\norm}[1]{\left \lVert #1 \right \rVert} 

\iccvfinalcopy 

\begin{document}

\author{
Stavros Tsogkas, Sven Dickinson\\
University of Toronto\\
27 King's College Circle Toronto, Ontario M5S 1A1 Canada\\
{\tt\small \{tsogkas,sven\}@cs.toronto.edu}
}

\title{AMAT: Medial Axis Transform for Natural Images}
\maketitle
\thispagestyle{empty}

\begin{abstract}
We introduce Appearance-MAT (AMAT), a generalization of the medial axis
transform for natural images, that is framed as a weighted geometric set cover problem.
We make the following contributions: 
i) we extend previous medial point detection methods for color images,
by associating each medial point with a local scale; 
ii) inspired by the invertibility property of the binary MAT, 
we also associate each medial point with a local encoding
that allows us to invert the AMAT, reconstructing the input image; 
iii) we describe a clustering scheme that takes advantage of the additional 
scale and appearance information to group individual points into medial branches, 
providing a shape decomposition of the underlying image regions.
In our experiments, we show \sota\ performance in medial point detection on
Berkeley Medial AXes (BMAX500), a new dataset of medial axes based on the BSDS500 
database, and good generalization on the SK506 and WH-SYMMAX datasets.
We also measure the quality of reconstructed images from BMAX500,
obtained by inverting their computed AMAT.
Our approach delivers significantly better reconstruction quality \wrt
three baselines, using just 10\% of the image pixels. 
Our code and annotations are available at~\url{https://github.com/tsogkas/amat} .
\end{abstract}

\section{Introduction}\label{sec:introduction}
\begin{figure}[!t]
    \centering
    \def\imageWidth{0.49}
    \subfloat[Input image]{\includegraphics[width=\imageWidth\linewidth]{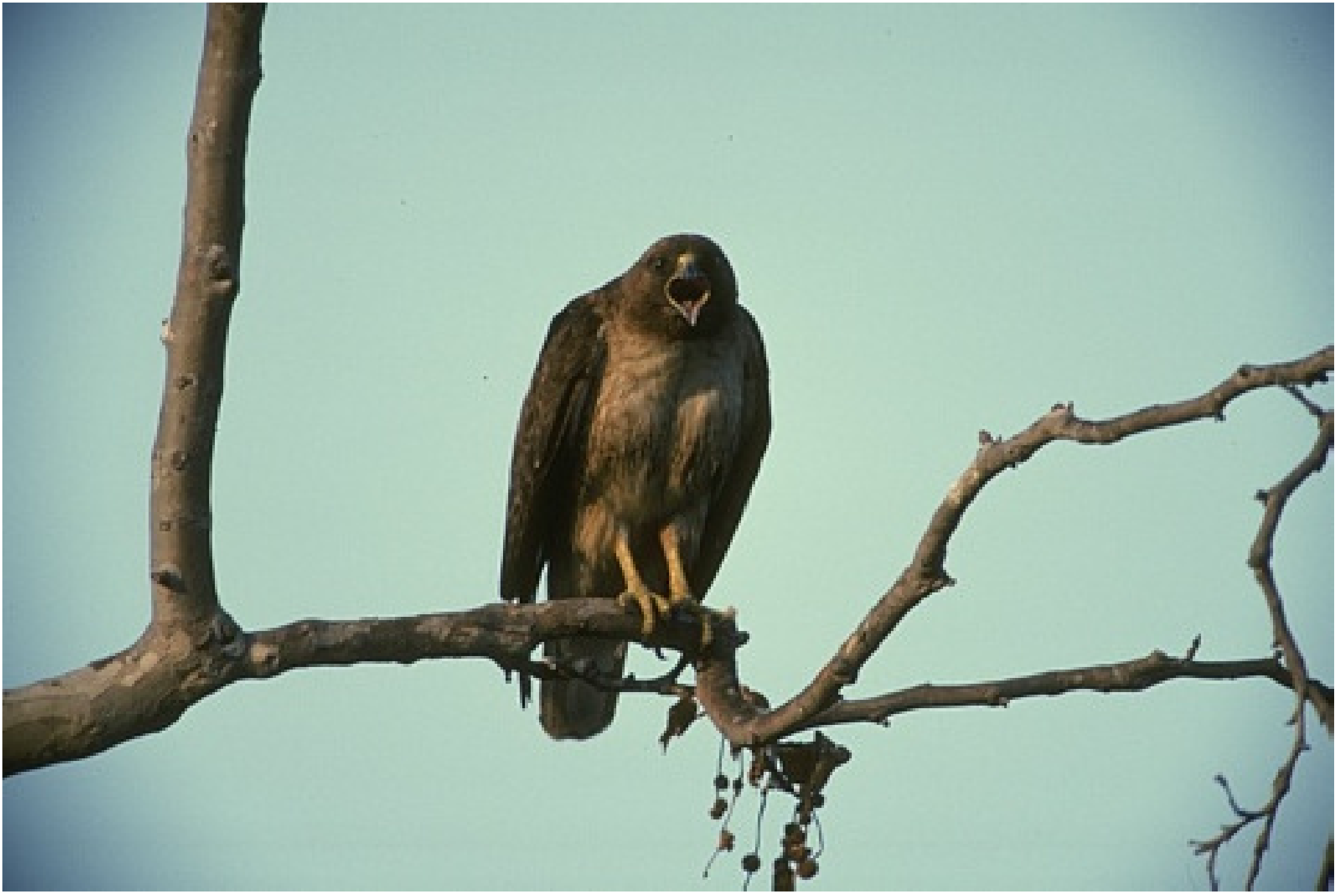}\label{fig:teaser:input}}
    \subfloat[Binary MAT]{\includegraphics[width=\imageWidth\linewidth]{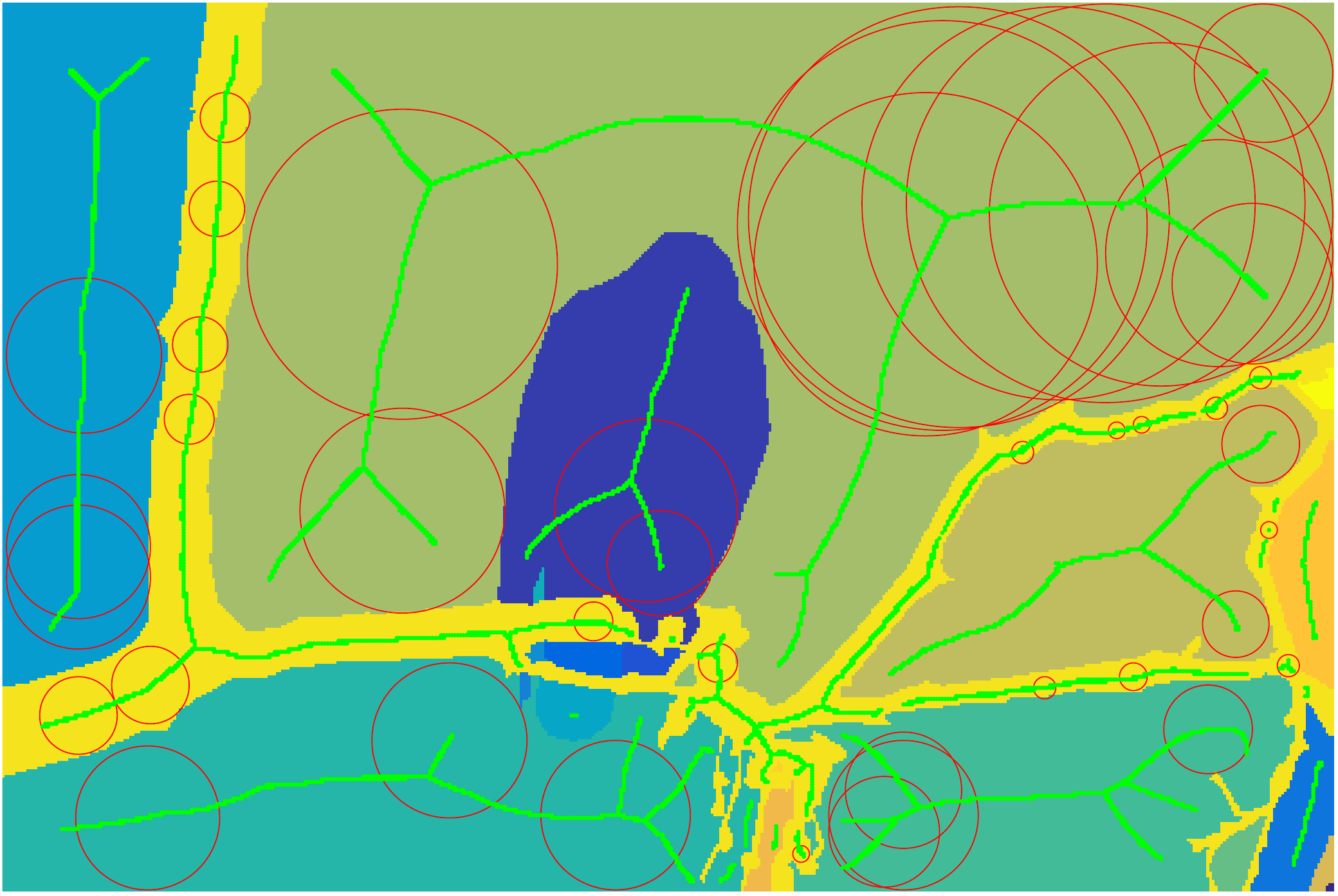}\label{fig:teaser:mat}} \\
    \subfloat[Appearance-MAT]{\includegraphics[width=\imageWidth\linewidth]{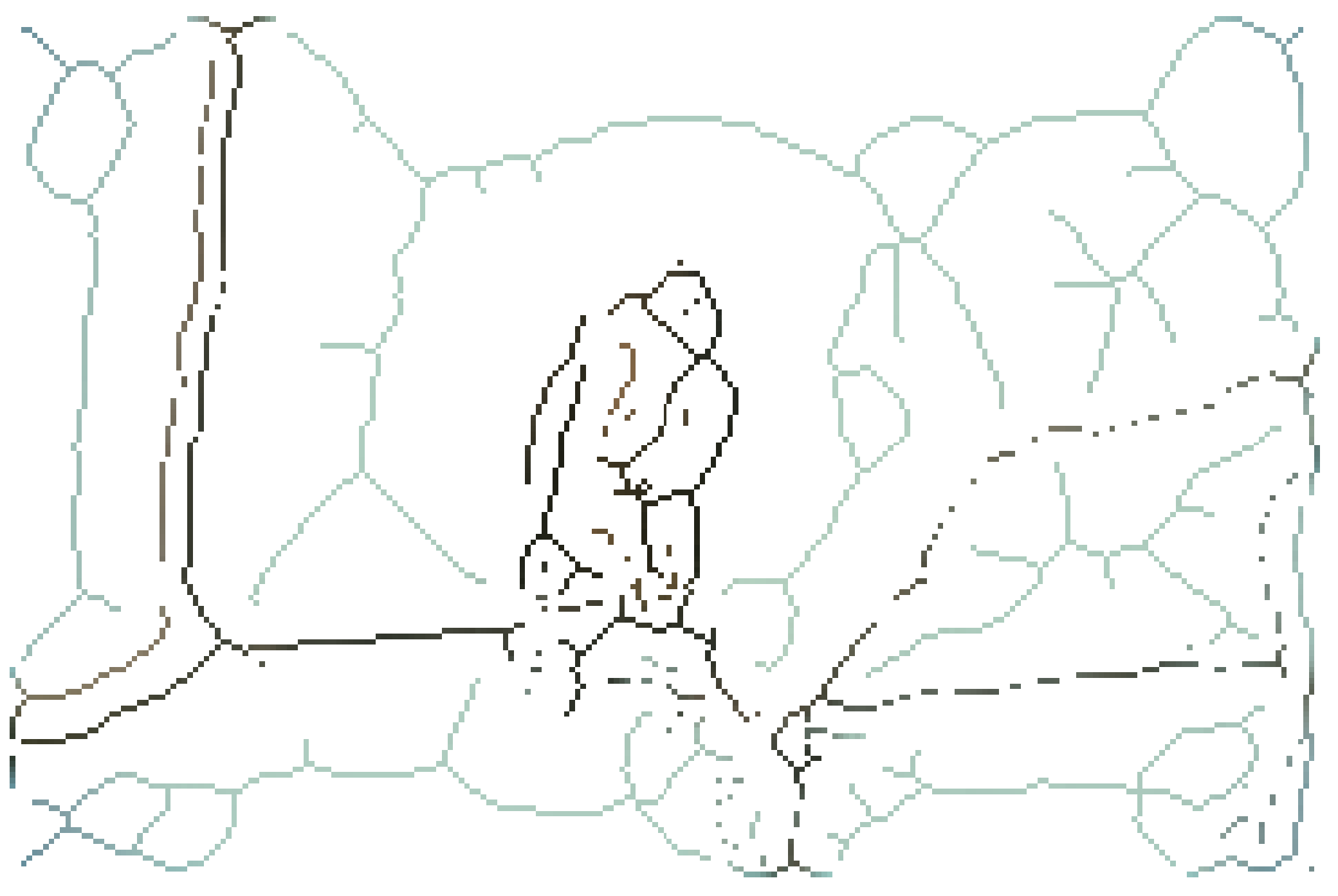}\label{fig:teaser:amat}}
    \subfloat[Reconstructed image]{\includegraphics[width=\imageWidth\linewidth]{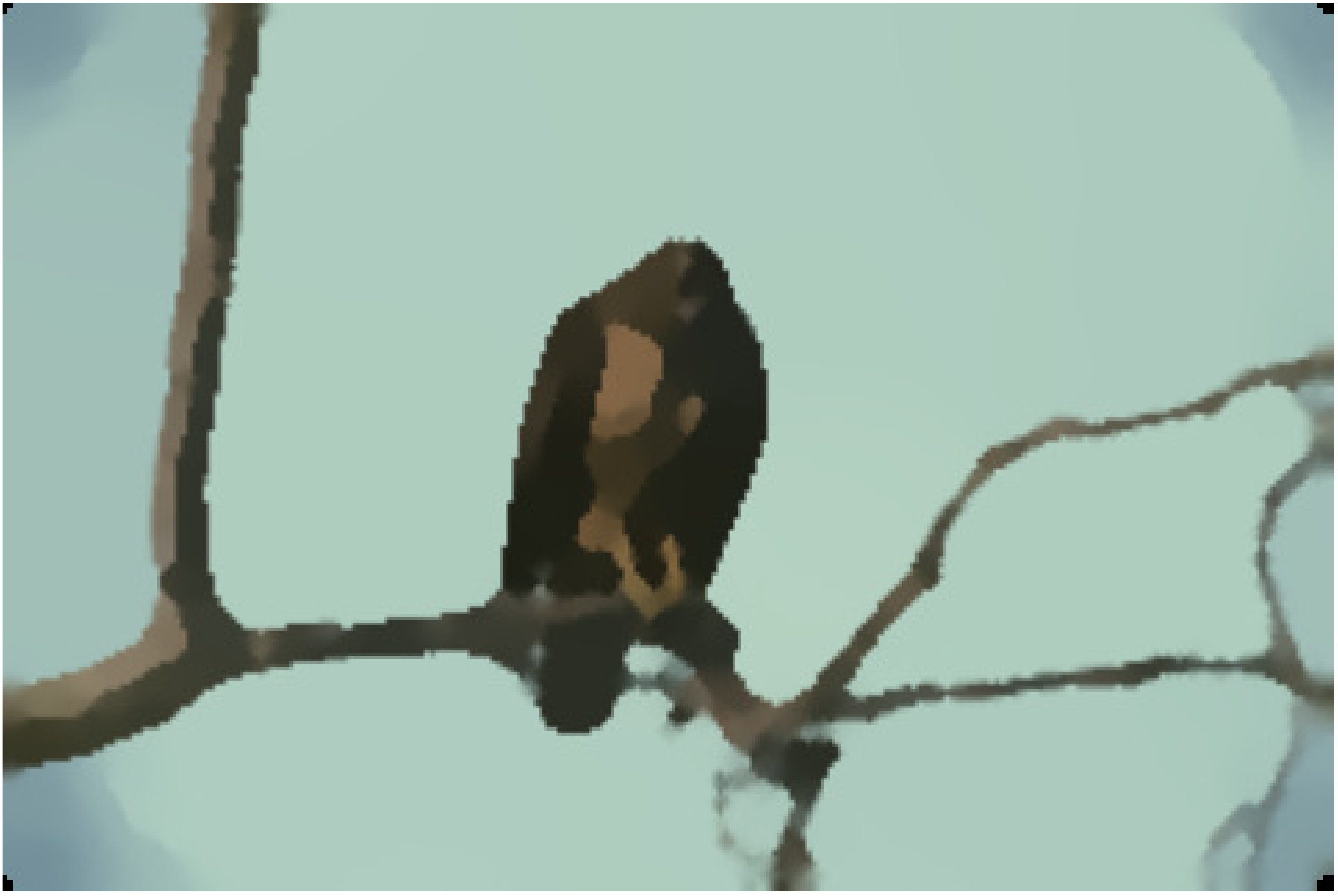}\label{fig:teaser:recon}}
    \caption{
        \textbf{Top:} Input image (\ref{fig:teaser:input}) and segmentation 
        (\ref{fig:teaser:mat}) from BSDS500,
        with color-coded ground-truth segments. 
        Medial axes (green) and a subset of medial disks (red) are overlaid. 
        Each (binary) segment can be reconstructed from its medial points and radii. 
        \textbf{Bottom:} Similarly, the AMAT (\ref{fig:teaser:amat}) carries enough information to reconstruct the 
        \emph{input image} (\ref{fig:teaser:recon}) with just $\sim 5\%$ of the pixels. 
    }
    \label{fig:teaser}
\end{figure}

Symmetry is a ubiquitous property in the natural world, with a well-established role in human vision.
Humans instinctively recognize and use symmetry to analyze complex scenes, as it facilitates the encoding of shapes and their discrimination and recall from memory~\cite{barlow1979versatility,royer1981detection,wagemans1998parallel}.
In the context of computer vision, \emph{local} symmetry is of particular interest, 
because of its robustness to viewpoint changes and its connection to salient structures, such as object parts.
This intuition is fundamental to many milestones in object representation theory, including generalized
cylinders~\cite{binford1971visual}, superquadrics~\cite{barr1981superquadrics}, 
geons~\cite{biederman1987recognition}, and shock graphs~\cite{siddiqi1999shock}.

Fundamental notions of local symmetry were introduced decades ago by Blum in the context 
of binary shapes with the \emph{medial axis transform (MAT)}~\cite{blum1967transformation,blum1973biological}.
The MAT is a powerful shape abstraction, and provides a compact representation that preserves topological
properties of the input shape. 
These properties are invariant to translation, rotation, scaling, articulation, 
and their locality offers robustness to occlusion.
The MAT has been very effective in reducing the computational complexity of algorithms for various tasks,
including shape matching~\cite{siddiqi1999shock} and recognition~\cite{sebastian2001recognition},
mesh editing~\cite{li2001decomposing,yoshizawa2003free}, and shape manipulation~\cite{du2004medial}.
For these reasons many researchers have tried to achieve a good balance
between MAT sparsity and reconstruction quality~\cite{tam2003shape,li2015q}.

Extending the notion of the MAT to natural images can correspondingly benefit applications that rely on
a sparse set of highly informative keypoints/landmarks, such as registration~\cite{zhou2016estimating}, 
retrieval~\cite{sivic2003video,avrithis2011medial}, pose estimation and body tracking~\cite{shotton2013efficient},
and structure from motion~\cite{agarwal2011building}.
It could also assist segmentation by enforcing region-based constraints through their medial point representatives
~\cite{teo2015detection}, and by providing a practical alternative to manual 
scribbles/seeds for interactive segmentation~\cite{boykov2001interactive,price2010geodesic,isack2016hedgehog,lin2016scribblesup}.
Another interesting application is artistic rendering of images:~\cite{gooch2002artistic} use approximate medial axes
to simulate brush strokes and generate a painting-like version of the input photograph.

Unfortunately, the MAT has not found widespread use in tasks 
involving natural images, due to the lack of a generalization that accommodates color and texture.
Previous works have mostly attacked \emph{medial point detection}~\cite{tsogkas2012learning,shen2016object},
which amounts to determining the \emph{locations} of points lying on medial axes 
but not the scale of the respective medial disks.
The type of axes considered is also typically constrained to make the 
problem more concrete:
\cite{tsogkas2012learning} only considers elongated structures, on either 
foreground objects or background; 
~\cite{shen2016object} focuses on \emph{object skeletons}, ignoring background structures.
These methods lack another key characteristic of the MAT: medial point locations alone do not provide sufficient
information to reconstruct the input.

In this paper we introduce the first ``complete'' MAT for natural images, dubbed \emph{Appearance-MAT (AMAT)}.
First, we provide a new definition in the context of natural images by framing MAT 
as a weighted geometric set cover (WGSC) problem.
Our definition is centered around the MAT invertibility property and elicits 
a straightforward criterion for quality assessment, in terms of the reconstruction of the input image.
Second, our algorithm associates each medial point with \emph{scale} as well as local \emph{appearance} information
that can be used to reconstruct the input.
Thus, the AMAT encompasses all the fundamental features of its  binary counterpart. 
Third, we describe a simple bottom-up grouping scheme that exploits the additional scale and appearance information to connect
points into medial \emph{branches}.
These branches correspond to meaningful image regions, and extracting them can support image segmentation
and object proposal generation, while offering a shape decomposition of the underlying structure as well. 

Being bottom-up in nature, our method does not assume object-level knowledge.
It computes medial axes of both foreground and background structures, 
yielding a compact representation that only uses $\sim 10\%$ of the image pixels.
Yet, this sparse set of points carries most of image signal,
differing from other sparse image descriptions, \eg edge maps, which strip the 
input of all appearance information.

We perform experiments in medial point detection on a new dataset of medial axes, the 
\emph{Berkeley-Medial AXes (BMAX500)}, which is built on the popular BSDS500 dataset, showing \sota\ performance.
We also measure the quality of reconstructions  obtained by inverting the AMAT of images from the same dataset, 
using  a variety of standard image quality metrics.
We compare with three reconstruction baselines: one built on the medial point detection algorithm from~\cite{tsogkas2012learning}
and two built from the ground-truth segmentations in BSDS500.
Our method significantly outperforms the baselines in terms of reconstruction quality, while attaining a $11\times$ compression ratio.

The outline of the paper is as follows: we start by reviewing related work on medial axis extraction for binary shapes
and natural images in~\refsec{sec:related}.
In~\refsec{sec:method} we describe our approach.
\refsec{sec:implementation} includes implementation details and in~\refsec{sec:experiments} we present our results.
Finally, in~\refsec{sec:discussion} we conclude and discuss ideas for future directions.

\section{Related Work}\label{sec:related}

\paragraph{Binary shapes:}\label{sec:related:binary}
Blum introduced the medial axis transform, or skeleton, of 2D shapes
in his seminal works~\cite{blum1967transformation,blum1973biological}.
Since then, researchers have developed algorithms for reliable
and efficient medial axis extraction, its extension to 3D shapes, and its application
to computer vision tasks.

Siddiqi~\etal define \emph{shocks} as the singularities of a curve evolution process acting on the boundaries of
a shape, and they organize them into a directed, acyclic shock graph~\cite{siddiqi1999shock}.
Shock graphs were successfully used in shape matching~\cite{siddiqi1999shock}, recognition~\cite{sebastian2001recognition},
and database indexing~\cite{sebastian2002shock}.
\emph{Bone graphs}~\cite{macrini2011bone} offer improved stability and a more intuitive representation of an object's parts, 
by identifying and analyzing ligature structures.
Visual part correspondences are also established and used to measure part and aggregated shape similarity in~\cite{latecki2000shape}.
The correspondence of skeleton branches to object parts is further explored in~\cite{ling2007shape,bai2008path}.
More recently, Stolpner \etal deal with the problem of approximating a 3D solid via a union
of overlapping spheres~\cite{stolpner2012medial}.

The value of the MAT has been equally appreciated by the graphics community, where object shapes 
are routinely represented as point clouds or triangular meshes.
Giesen~\etal~\cite{giesen2009scale} introduced the \emph{scale axis transform}, a skeletal shape representation
that yields a hierarchy of successively simplified skeletons, which are obtained by multiplicative scaling of the
MAT's radii.
Li~\etal~\cite{li2015q} use quadratic error minimization to compute an accurate linear approximation of the MAT, called \emph{Q-MAT}.
They show experiments on medial axis simplification where they reduce the number of nodes of an initial medial mesh
by three orders of magnitude, while preserving good surface reconstruction.
A comprehensive compilation of medial methods and their applications in the binary setting can be found in~\cite{siddiqi2008medial}.

\paragraph{Natural images:}\label{sec:related:natural}
Compared to the binary setting, the number of works on medial axis detection for natural images is rather limited.
Levinstein \etal~\cite{levinshtein2013multiscale} detect \emph{symmetric parts} of objects
by learning to merge adjacent deformable, maximally inscribed disks, modeled as superpixels.
Learned attachment relations are then used to combine detected parts into coarse skeletal representations.
Lee \etal extend that work by introducing a deformable disk model that can capture curved and tapered parts, and also add
continuity constraints to the medial point grouping process~\cite{lee2013detecting}.
In other works medial point detection is posed as a classification problem where pixels are labeled
as ``medial'' or ``not-medial'', inspired by similar methods for boundary detection~\cite{martin2004learning}.
Tsogkas and Kokkinos use multiple instance learning (MIL) to deal with the unknown scale and orientation 
during training~\cite{tsogkas2012learning}, while Shen \etal adapt a CNN with 
side outputs~\cite{xie2015holistically} for object skeleton extraction~\cite{shen2016object}.
All these approaches exploit appearance information by incorporating a machine learning algorithm.

Our work can be regarded as lying at the intersection of previous work on binary and natural images.
From a technical standpoint, it shares more similarities with binary methods, for instance~\cite{stolpner2012medial},
which solves the set cover problem for volumes in the 3D space.
At the same time, it can be applied to real images, without assuming a figure-ground segmentation,
but it also demonstrates unique characteristics.
Our method does not involve learning, and is not constrained in detecting a particular subset of
medial axes as~\cite{tsogkas2012learning,shen2016object}.
It also complements existing methods by augmenting point locations with scale and appearance descriptions, which
are necessary for reconstructing the input.

\section{AMAT definition}\label{sec:method}

\begin{figure*}[t]
	\centering
	\includegraphics[width=0.3\linewidth]{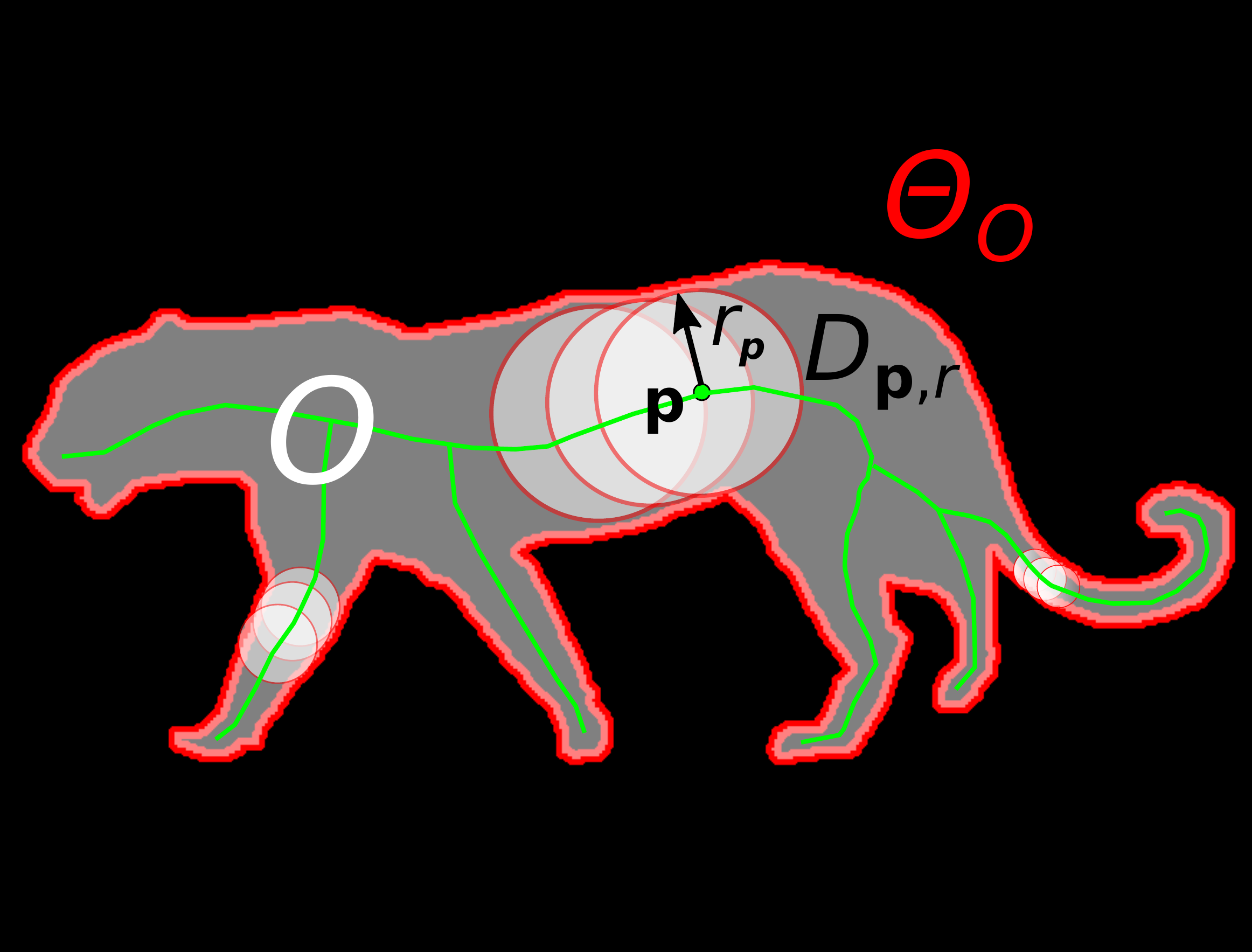}\hfill
	\includegraphics[width=0.35\linewidth]{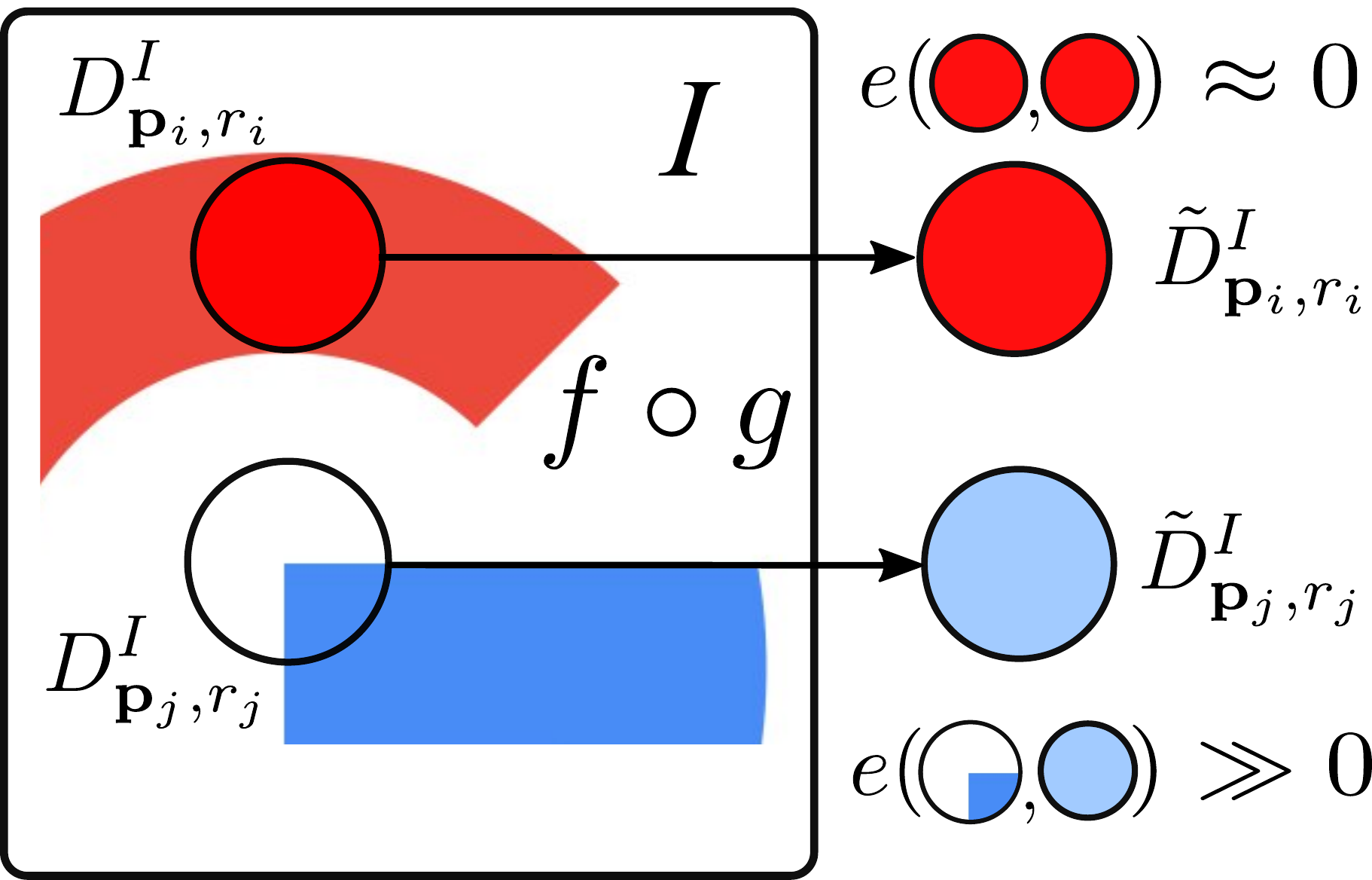}\hfill
	\includegraphics[width=0.22\linewidth]{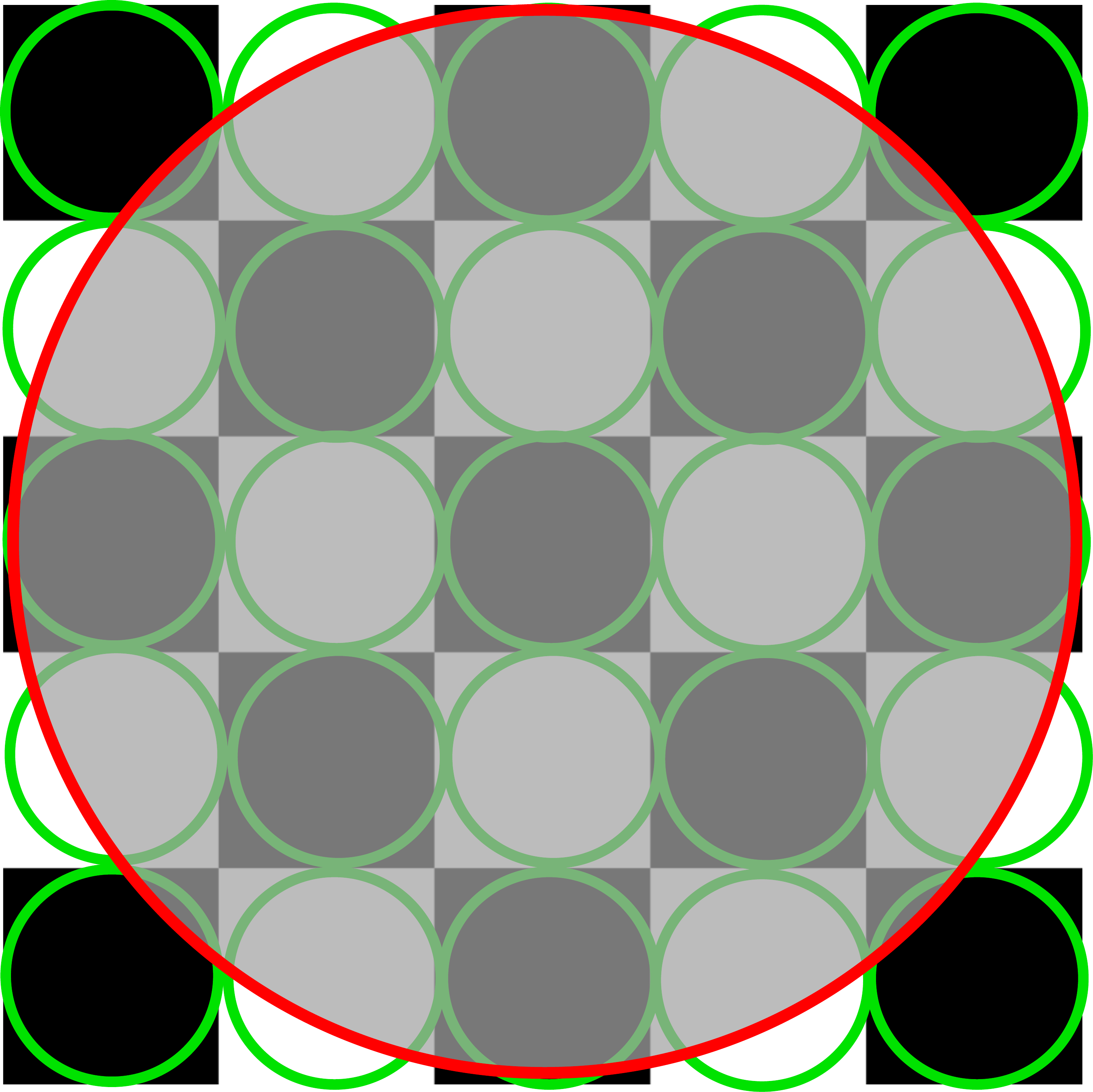}
	\caption{
		\textbf{Left:} We can reconstruct a binary shape by expanding a value 
		of ``1'' within the area of all medial disks.
		\textbf{Middle:} Disks are represented by their mean RGB value; 
		disks that cross region boundaries have a high reconstruction error.
		\textbf{Right:} Toy example: depending on the task, the user can favor a dense 
		representation with low reconstruction error (green disks) or a sparse 
		representation with high reconstruction error (red disk) by varying the scale parameter $w_s$.   
	}
	\label{fig:method:definition}
\end{figure*}


Consider a 2D binary shape, $O$, like the one in~\reffig{fig:method:definition}, and its boundary $\Theta_O$.
The \emph{medial axis} of $O$ is the set of points $\p{}$ that
are centers of the maximally inscribed (medial) disks, bitangent to $\Theta_O$
in the interior of the shape. The \emph{medial} (disk) \emph{radius} $r_{\p{}} \equiv r(\p{})$
is the distance between $\p{}$ and the points where the disk touches $\Theta_O$.
The process of mapping $O$ to the set of pairs $(\p{},r_{\p{}}) \in \mathbb{R}^2 \times \mathbb{R}$
is called the \emph{medial axis transform} (MAT).
Given these pairs, we can reconstruct $O$ as a union of overlapping disks that sweep-out 
its interior by ``expanding'' a value of one (1) inside the area covered by each  medial disk.

We argue that a MAT for real images should satisfy a similar principle: given the MAT of an image, 
we should be able to ``invert'' it, reconstructing \emph{the image} itself.
There are several reasons why extending this idea to real images is a challenging task: 
natural images depict complex scenes, cluttered with numerous objects, instead of 
just a single foreground shape.
Moreover, unlike binary images, real images exhibit complicated color and texture distributions.
Nevertheless, we can exploit image redundancies and assume that an image is 
composed of many small regions of relatively uniform appearance.
This is the same assumption that underlies most superpixel algorithms which 
break up an image into non-overlapping patches, 
while respecting perceptually meaningful region 
boundaries~\cite{shi2000normalized,levinshtein2009turbopixels,achanta2012slic}. 

\paragraph{Notation.} 
In the rest of the paper we denote a disk of radius $r$, 
centered at point $\p{}$, as $D_{\p{},r}\equiv D(\p{},r)$. 
For brevity, we often refer to such a disk as a $r$-disk or $(\p{},r)$-disk.
$\set{D}$ is a collection of such disks of varying centers and radii, 
$\set{D}=\{ D_{\p{i},r_{\p{i}}} \},\, i\in\mathbb{N}$.
The intersection of a $(\p{},r)$-disk with an image $I$ is a disk-shaped region 
of the image, and is denoted by 
$I \cap D_{\p{},r}=D_{\p{},r}^I \subset \set{D^I} = \{ D^I_{\p{i},r_{\p{i}}} \}$. 
Finally, we use $\circ$ to denote function composition, and $\norm{\cdot}$ for 
an appropriate error metric (\eg, the $L_2$ norm).

\paragraph{Formulation.} 
Consider an RGB image $I\subset\R^3$, and a disk-shaped region $D_{\p{},r}^I \subset I$.
Let $f:\set{D^I} \rightarrow \R^K$ be a function that maps $D_{\p{},r}^I$ to a vector  $\f{\p{},r}=f\circ D_{\p{},r}^I$; 
we call $\f{\p{},r}$ the \emph{encoding} of $D_{\p{},r}^I$. 
Now let $g:\R^K \rightarrow \set{D^I}$ be a function that maps $\f{\p{},r}$ back to a disk patch 
$\tilde{D}_{\p{},r}^I = \g{\p{},r} = g \circ \f{\p{},r}$. We call $g$ the \emph{decoding} function.
In the general case, $f$ and $g$ will be \emph{lossy} mappings, which means that the reconstruction error 
$e_{\p{},r} = \norm{ \tilde{D}_{\p{},r}^I - D_{\p{},r}^I} \geq 0$. 
Using the above, we define the AMAT as the set of tuples 
$M:\{ ( \p{1},r_{\p{1}}, \f{\p{1},r_{\p{1}}} ), \ldots, ( \p{m},r_{\p{m}}, \f{\p{m},r_{\p{m}}} ) \}$, such that:
\begin{equation}
	M = \argmin_{\p{},r}{\sum_{i=1}^m e_{\p{i},r_i}},\quad I=\bigcup_{i=1}^m D_{\p{i},r_i}^I.
	\label{eq:minimization}
\end{equation}
In~\refsec{sec:method:wgsc} we discuss constraining $m$.

\paragraph{Encoding and decoding functions.}
Our framework allows $f,g$ to take any form; 
for example, $f$ could be a histogram representation of color in $D_{\p{},r}^I$ and $g$ could return the mode of the distribution.
In this paper we opt for simplicity:
$f$ computes the mean of each color channel ``summarizing'' $D_{\p{},r}^I$, in a $3\times1$ vector $\f{\p{},r}$.
Conversely, $g$ constructs an approximation $\tilde{D}_{\p{},r}^I \approx D_{\p{},r}^I$ by replicating $\f{\p{},r}$ in the
respective disk-shaped area.
When the $(\p{},r)$-disk is fully enclosed in a uniform region the reconstruction error $e_{\p{},r}$
is low, whereas when the disk crosses a strong image boundary, the encoding $\f{\p{},r}$ cannot accurately represent
the underlying image region, resulting in a higher error. 

Note that the definition in~\refeq{eq:minimization} suggests conceptual similarities with superpixel representations.
Selecting the points $\{ (\p{i},r_i,\f{\p{i},r_i}) \},\,i=1,\ldots,m$ is equivalent to covering the input image
with $m$ disk-shaped superpixels. Minimizing the total reconstruction error implies that these ``superdisks'' do not
cross region boundaries, as this would incur a high reconstruction error, as shown in~\reffig{fig:method:definition}.
However, there are two important differences:
First, in our case a canonical shape (disk) is used, whereas superpixels can have any form. 
Second, our disks are \emph{overlapping}, in contrast to standard, non-overlapping superpixels.

Using canonical shapes helps achieve sparsity of the final MAT. Disks are optimal in that sense, as they are rotationally invariant and are fully defined using only their center and radius. By contrast, a free-form element requires storing coordinates of \emph{all} its boundary points. 
On the other hand, using one shape and no overlap would not reduce reconstruction 
quality, but it would result in disjointed medial points instead of smooth, connected medial axes.

\subsection{AMAT as a Geometric Set Cover Problem}\label{sec:method:wgsc}
The geometric set cover is the extension of the well studied set cover problem, in a geometric space.
Here we only consider the case of a two-dimensional space and we particularly focus on the 
\emph{weighted} version of the problem, which is defined as follows:
Consider a universe of $N$ points $X \in \R^2$ and subsets
$\set{D} = \{D_1,D_2,\ldots,D_k\} \subseteq X$, called \emph{ranges}. 
A common choice for $D_i$ is intersections of $X$ with simple shape primitives, such as disks or rectangles.

Now assume that each element in $\set{D}$ is associated with a non-negative weight or \emph{cost} $c_i$.
Solving the WGSC problem amounts to finding a sub-collection $\bar{\set{D}} \subset \set{D}$ that covers the entire $X$
(all $N$ elements of $X$ are contained in at least one set in $\bar{\set{D}}$), while having the minimum
total cost $C$; the total cost is simply the sum of costs of individual elements in $\bar{\set{D}}$.
WGSC is a strongly NP-hard problem for which polynomial-time approximate solutions (PTAS) exist.
The interested reader can find more details on WGSC and related algorithms 
in~\cite{mustafa2015quasi,varadarajan2010weighted,har2012weighted,chan2012weighted}.

The AMAT formulation lends itself naturally to a WGSC interpretation.
The spatial support $X^I$ of an input image $I$, is the universe of $N$ points.
As $\set{D}$ we consider the set of $r$-disks with $r$ chosen from a finite set $\set{R}:\{r_1,r_2,\ldots,r_R\}$.
The $r$-disks can be placed at any position $\p{}=(x,y)\in X^I$ such that 
$D_{\p{},r}$ is fully contained in $X^I$.
We also assign a cost $c_{ij} \equiv c_{\p{i},r_j} \propto e_{ij}$ to each 
$(\p{i},r_j)$-disk, $i\in[1,N],\, j\in[1,R]$.
Note that for brevity, we drop the subscripts $\p{i},r_j$ and simply use $ij$.
We provide more details regarding computation of $c_{ij}$ in~\refsec{sec:implementation:diskcost}.

As~\refeq{eq:minimization} suggests, the goal is to find a subset of disks that cover the entire image, while maintaining
a low total reconstruction cost. 
A trivial solution would be to select each pixel as a disk of radius $r=1$, in which case
$M=\{ (\p{1},r_{\p{1}},f_{\p{1},r_{\p{1}}}), \ldots, ( \p{N},r_{\p{N}},f_{\p{N},r_{\p{N}}} ) \}$,
and $\sum_{i=1}^N e_{\p{i},r_i} = 0$; each pixel can be perfectly represented by its mean value.
Such a solution is of no practical use. 
Staying true to the spirit of the MAT, we seek a solution that is \emph{sparse}
(low number of medial points $m$), while being able to adequately reconstruct the input image.
One possible way to do this would be to agree on a fixed ``budget'' of points, and look for the 
optimal solution, given $m$.
However, choosing an acceptable $m$ can be a nuisance, as its value can vary significantly  from image to image.

In the original MAT, sparsity is implicitly induced through the use of maximal disks,
touching the shape boundary at two or more  points.
Extending the maximality principle to real images is not straightforward
because color and texture boundaries are not robustly defined.
Relying on the output of an edge extraction algorithm is not a viable option either,
as it would make our method sensitive to errors from which it would be impossible to recover.

\begin{figure*}[t]
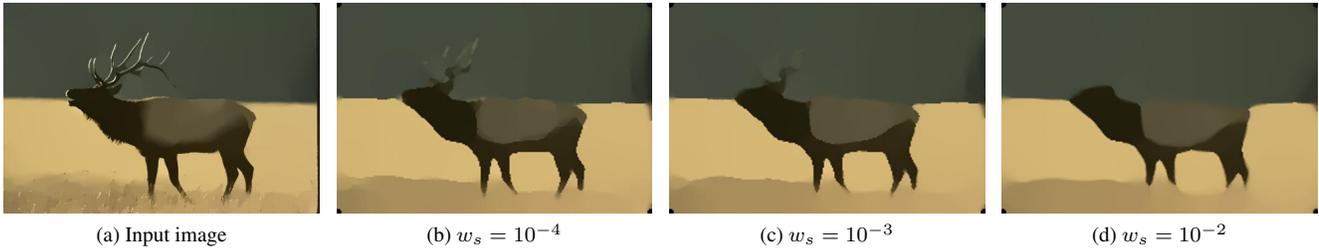

\def\img_id{41004}
\def\imageWidth{0.24}
\subfloat[Input image]{\includegraphics[width=\imageWidth\textwidth]{\img_id_smoothed.jpg}}\hfill
\subfloat[$w_s=10^{-4}$]{\includegraphics[width=\imageWidth\textwidth]{{\img_id_recon0.0001}.jpg}}\hfill
\subfloat[$w_s=10^{-3}$]{\includegraphics[width=\imageWidth\textwidth]{{\img_id_recon0.001}.jpg}}\hfill
\subfloat[$w_s=10^{-2}$]{\includegraphics[width=\imageWidth\textwidth]{{\img_id_recon0.01}.jpg}}\hfill
\caption{Using a progressively larger scale-cost factor $w_s$ removes details, keeping only coarse image structures.}
\label{fig:smoothing}
\end{figure*}

Instead, we choose to regularize the minimization criterion 
in~\refeq{eq:minimization} by adding a scale-dependent term 
$s_j = \frac{w_s}{r_j} \propto \frac{1}{r_j}$ to the costs $c_{ij}$. 
This way we favor the selection of larger disks at each point, as long as $s_j$ 
is not ``too'' large with respect to the error incurred by picking
$D_{\p{},r_{j+1}}$ instead of $D_{\p{},r_j}$.
Selecting a high value for $w_s$ leads to a sparser solution with higher total reconstruction error, whereas a low value for $w_s$ aims for a better
reconstruction, by utilizing more, smaller disks to cover $X^I$.
\reffig{fig:method:definition} (right) shows a toy example of these two cases and
\reffig{fig:smoothing} shows how varying $w_s$ progressively removes details
in a real image, keeping only the coarser structures.

\paragraph{Greedy approximation algorithm.}
There are many polynomial-time-approximate-solution (PTAS) algorithms for the vanilla set cover problem
and its geometric variants.
Here we use the simple, greedy algorithm described in~\cite{vazirani2013approximation}, adapted for the weighted case.
The steps of our method are described in~\refalg{alg:greedy}.
\begin{algorithm}[t]
	\caption{AMAT greedy algorithm.}
	\label{alg:greedy}
	\begin{algorithmic}[1]
		\Statex \textbf{Input:} $X^I=\{\p{1},\ldots,\p{N}\},\set{R}=\{r_1,\ldots,r_R\},f,g$
		\Statex \textbf{Output:} $M$
		\State Initialization: $M \leftarrow \emptyset,X^c \leftarrow \emptyset$ \Comment{$X^c:$ covered pixels}.
		\State Compute $\f{\p{},r},\, \g{\p{},r} = g \circ \f{\p{},r},c_{\p{},r},\  \forall \p{} \in I, \forall r \in \set{R}$
		\While{$X^c \subset X^I$}
			\State $c_{\p{},r}^e \leftarrow \frac{c_{\p{},r}}{|D_{\p{},r} \setminus X^c|}+\frac{w_s}{r},\ \forall \p{} \in X^I,\ \forall r \in \set{R}$
			\State $(\p{}^{*},r^{*}) \leftarrow \argmin_{\p{},r}{c_{\p{},r}^e},$		
			\State $c_{\p{},r} \leftarrow c_{\p{},r} - \frac{c_{\p{},r}}{|D_{\p{}^{*},r^{*}} \setminus X^c|},$
			\Statex\hspace{\algorithmicindent}\hspace{\algorithmicindent}
				$\forall \p{},r: D_{\p{}^{*},r^{*}} \cap D_{\p{},r} \neq \emptyset$
			\State $M\leftarrow M\cup{(\p{}^{*},r^{*},f_{\p{}^{*},r^{*}})}$
			\State $X^c \leftarrow X^c \cup D_{\p{}^{*},r^{*}}$ 
		\EndWhile
	\end{algorithmic}
\end{algorithm}
We start by computing the costs $c_{ij}$ for all possible disks $D_{ij}$.
We define the \emph{effective cost} of $D_{ij}$ as $c_{ij}^e = \frac{c_{ij}}{A_{ij}} + s_j$ , where $A_{ij}$ is the number
of \emph{new} pixels covered by $D_{ij}$ (pixels that have not been covered by a previously selected disk).
Starting from an empty set $M$, we pick the disk with the lowest $c_{ij}^e$ and add it to the solution, 
removing the area $D_{ij}$ from the remaining pixels to be covered.
We also adjust the cost of all disks that intersect with $D_{ij}$, 
because each disk should be penalized only for the \emph{new} pixels it is covering.
This process is repeated until all image pixels have been covered by at least one disk.

\subsection{Grouping Medial Points Into Branches}\label{sec:method:grouping}
The scale and appearance associated with each medial point provide a rich
description that can be used to group points belonging to the same region into \emph{medial branches}.
The beneficial effects of grouping in low-level vision tasks have been
observed in previous works~\cite{felzenszwalb2006min,zhu2007untangling,kokkinos2010highly,qi2015making}.
In our case, grouping pixels into branches can help us refine the final medial axis, 
by aggregating consensus from neighboring points, and break the image into meaningful regions.

We group detected medial points using an agglomerative scheme that starts at fine scales and
progressively merges together nearby points at coarser scales.
Our grouping criterion relies on proximity in \emph{scale-space} and \emph{appearance}.
Intuitively, points that lie close have higher probability of belonging to the same branch.
We also expect that the scale of points will change \emph{gradually} along a branch,
so points that lie close to each other but have very different radii should probably not be grouped together.
Finally, two points should not be grouped if their encodings are very dissimilar,
regardless of their proximity in scale-space. 

We initialize branches as the connected components of the AMAT output.
Starting at a scale $r_j$, we consider one branch at a time, and examine all other
branches within a neighborhood of size $r_j \times r_j$ and a scale neighborhood $[r_{j-3},r_j]$.
If two branches coexist in this scale-space neighborhood and their average encodings 
(summed along the branch curve) are similar, they are merged.
The grouping algorithm terminates when all scales have been considered.


\subsection{Medial Branch Simplification}\label{sec:method:simplification}
The output of our algorithm captures mostly region centerlines but there are still
imperfections in the form of noisy, disconnected medial point responses 
or ``lumps'', instead of thin contours.
Such imperfections are expected because of the approximate solution
to the minimization problem of~\refeq{eq:minimization} and the use of a discrete grid.

Grouping MAT points into branches makes it possible to process each branch individually, enabling
the correction of these errors post hoc.
We perform simple morphological operations (dilation and thinning) 
on the points of each branch to merge neighboring and isolated pixels together, while removing 
redundant responses. 
We also adjust the scales of the medial points, to ensure that the medial disks corresponding 
to the simplified structure span the same image area.
Because grouped branches correspond to relatively homogeneous regions, reconstruction
results after simplification are practically identical.
Examples of simplified medial axes are illustrated in~\reffig{fig:experiments:detection}.

\section{Implementation Details}\label{sec:implementation}

\paragraph{Disk Cost Computation.}\label{sec:implementation:diskcost}
Using a simple error metric such as MSE to compute $c_{ij}$ is not effective since 
disks with low MSE scores do not necessarily respect image boundaries.
We propose the following alternative heuristic:
First, we convert the RGB image to the CIELAB color space which is more suitable 
for measuring perceptual distances.
Then, we define the cost of $D_{ij}$ as
\begin{equation}
    c_{ij} = \sum_k \sum_l \norm{\f{ij} - \f{kl} }^2 \quad \forall k,l: D_{kl} \subset D_{ij}.
    \label{eq:diskcost}
\end{equation}
Intuitively, a low cost $c_{ij}$ implies that the encoding $\f{ij}$ is representative of \emph{all}
disks that are fully contained in $D_{ij}$, hence $D_{ij}$ is not crossing any region boundaries.

\paragraph{Dealing With Texture.}\label{sec:implementation:texture} 
The main motivation behind the choice of simple functions $f,g$,
was simplicity and  computational efficiency.
Such functions also allow us to inject certain desired characteristics in the 
AMAT solution, such as appearance uniformity and alignment with boundaries. 

However, natural images often contain high-frequency textures or noise, which 
can lead to the accumulation of large errors 
in~\refeq{eq:diskcost}, and promote the selection of disks that do not correspond to perceptually coherent regions. 
Simple processing techniques (\eg, Gaussian filtering) can reduce noise but they also degrade image boundaries and
blend together neighboring regions.

To alleviate this problem, we ``simplify'' the input image before extracting the AMAT, using a method that smooths high frequency
regions, while preserving important edges~\cite{xu2011image}.
In practice, this preprocessing produces an image that is  perceptually very similar to the original, 
but without high-frequency textures that can cause the greedy algorithm to fail by placing disks at undesired locations.

\paragraph{Inverting the AMAT.}\label{sec:implementation:inverting}
Generating the reconstruction of a single disk-shaped region, $\tilde{D}_{\p{},r}^I$, is trivially achieved by
replicating $\f{\p{},r}$.
However, since medial disks overlap, most pixels in the image domain will be covered by multiple disks with different encodings.
We resolve this ambiguity in a simple way: while computing the AMAT, we keep 
track of the  number of disks each pixel is covered by; this quantity is called 
\emph{depth} in the context of the set cover problem.
We then use the average $\f{}$ of all disks covering a point $\p{i}$ with depth $d_i$ as its reconstructed value:
\begin{equation}
    \tilde{I}(\p{i}) = \frac{1}{d_{\p{i}}} \sum_{\p{},r} \f{\p{},r}, \quad \forall \p{},r: \p{i}\in D_{\p{},r}.
    \label{eq:reconstruction}
\end{equation}

\paragraph{Parameter Values.}\label{sec:method:parameter}
For the smoothing algorithm we use the default values $\lambda=2\cdot10^{-4}$ and $\kappa=2$ that 
the authors suggest for natural images~\cite{xu2011image}.
Regarding the scale cost term described in~\refsec{sec:method:wgsc}, we found that $w_s=10^{-4}$ is a value that 
strikes a good balance between reconstruction quality and sparsity of the generated medial axis.
The maximum radius $R$ must be finite to keep complexity manageable, but large 
enough to capture large uniform structures in the image. 
Based on the size of images used in our experiments we used $40$ scales,  
excluding $r=1$ to force disks to be larger than single pixels; thus $r\in[2,41]$.

\paragraph{Complexity and Running Time.}\label{sec:method:complexity}
Computing $c_{ij}$ requires computing differences for all disks in $D_{ij}$.
If $r_j$ is large, this number can grow quickly, yielding $O(NR^4)$ complexity.
However, the most time-consuming step is the greedy approximation algorithm: 
At each iteration we cover at most $O(R^2)$ pixels, but we also have to
update the costs of all overlapping disks.
This has $O(NR^2\sum_{r=1}^R r^2) = O(NR^5)$ complexity.
One could parallelize the procedure by partitioning an image, simultaneously processing individual parts, and combining the results.
Our single-thread \matlab\ implementation takes $\sim \unit[30]{sec}$ for the AMAT, 
grouping, and simplification steps, on a $256\times256$ image.

\section{Experiments}\label{sec:experiments}
\begin{figure*}[t]
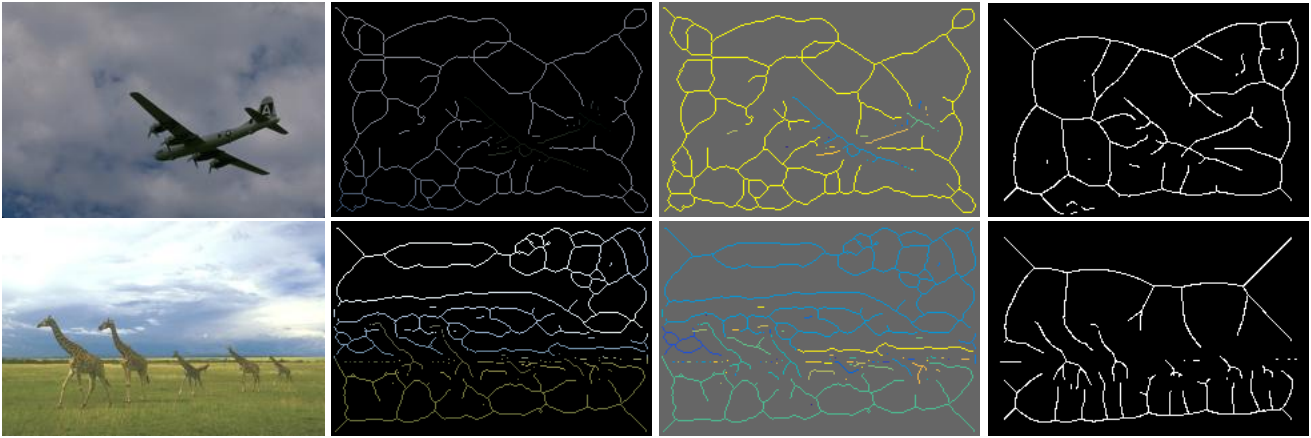

    \centering
    \def\imageWidth{0.245}
    \def\img_id{3096}
    \includegraphics[width=\imageWidth \textwidth]{\img_id_resized.jpg}
    \includegraphics[width=\imageWidth \textwidth]{\img_id_axes_simplified.pdf}
    \includegraphics[width=\imageWidth \textwidth]{\img_id_branches_simplified.pdf}
    \includegraphics[width=\imageWidth \textwidth]{\img_id_gt_skel.pdf}

    \def\img_id{253055}
    \includegraphics[width=\imageWidth \textwidth]{\img_id_resized.jpg}
    \includegraphics[width=\imageWidth \textwidth]{\img_id_axes_simplified.pdf}
    \includegraphics[width=\imageWidth \textwidth]{\img_id_branches_simplified.pdf}
    \includegraphics[width=\imageWidth \textwidth]{\img_id_gt_skel.pdf}

    \caption{
        From left to right: Input image, AMAT axes (unused points in black),
        medial point groups (color-coded), ground-truth skeletons.
        Note that semantically coherent image regions (\eg, sky, grass) tend to be grouped together. 
    }
    \label{fig:experiments:detection}
\end{figure*}

We evaluate the performance of our method on two tasks: 
i) localization of medial points in an image; and
ii) generating accurate reconstructions of images, given their AMAT.

\subsection{Medial Point Detection}\label{sec:experiments:detection}
We want to emphasize the difference between the problem we are addressing and the objectives pursued in other works.
In~\cite{tsogkas2012learning} the authors focus on detecting local reflective symmetries of elongated structures,
and they build a dataset with annotations of segments in the BSDS300 that fit this description.
As a result, a large portion of the segments in BSDS is not used in performance evaluation.
In~\cite{shen2016object} the authors are explicitly interested in extracting \emph{object}
skeletons, completely ignoring background structures.
Although extracting object skeletons may be convenient for some tasks, it does not constitute a generalized
notion of MAT.

In our work we do not make such distinctions. 
The central idea behind the AMAT is to be able to reproduce the full input image,
so we view all parts of the image as equally important.
This is also the reason we choose BSDS500 as a basis for constructing medial axes annotations.
BSDS500 contains multiple segmentations for each image, offering higher probability of
capturing segments at varying scales, making it more relevant to the problem we are trying to solve
than datasets with object-level annotations.

Following~\cite{tsogkas2012learning}, 
we individually apply a skeletonization algorithm~\cite{telea2002augmented} 
to binary masks of all segments in a given segmentation, extracting \emph{segment skeletons}.
The medial axis ground-truth for the image is formed by taking the union of all the segment skeletons, and this
process is repeated for all available annotations (usually 5-7 per image).
To conduct a fair comparison, we retrain the CG+BG+TG variant (MIL-color) from~\cite{tsogkas2012learning} on BMAX500.
We also tried to retrain the CNN used in~\cite{shen2016object}, but the outputs we obtained were too noisy, 
and of no practical use.
We hypothesize that this is because of the lack of consensus among the multiple ground-truth maps
available for each image, which leads to convergence problems for the network; this has been previously
reported in~\cite{xie2015holistically}.
We evaluate performance using the standard precision, recall and F-measure metrics, 
and show the superior results of our method in~\reftab{tab:experiments:medialpointdetection-BMAX500}.
Note that our algorithm outputs binary skeletons, so plotting a PR-curve
by varying a score threshold is not applicable in our case.
``Human'' performance is defined in the same manner as in~\cite{martin2004learning,tsogkas2012learning}.
For all methods, detections within a distance of $1\%$ of the image diagonal from a ground-truth positive 
are considered as true positives.
We show qualitative results of the medial axes and the grouped branches in~\reffig{fig:experiments:detection}.

\paragraph{Segmentation + skeletonization:} 
As an additional baseline we compute skeletons after running Arbelaez's
segmentation algorithm~\cite{arbelaez2006boundary,arbelaez2011contour} at 
scales 0.2 (F=0.61), 0.3 (F=0.58), 0.4 (F=0.54), 0.5 (F=0.5). 
We point out that the performance of UCM + skeletonization depends critically 
on the threshold selection. 
The optimal threshold is not known a-priori and, given a desired level of 
skeleton detail, the appropriate value varies from image to image. 
By contrast, AMAT's scale parameter is more intuitive to select and provides 
image-independent control of skeleton detail.

\paragraph{SK506 and WH-SYMMAX:}
We also evaluate the performance of the AMAT on two additional datasets: 
WH-SYMMAX~\cite{shen2016multiple} (F=0.44) and SK506~\cite{shen2016object} (F=0.33).
We compare with the pretrained FSDS~\cite{shen2016object} evaluating only on 
foreground skeletons, since our approach does not distinguish foreground from background.
FSDS performs better than AMAT (F=0.67 and F=0.45 respectively). 
This is unsurprising, given that FSDS is a supervised method trained on these 
datasets in a way that allows it to take advantage of rich, object-specific 
information. 
However, this specialization comes at a cost: FSDS cannot generalize well to 
structures it has not seen before, which is evident when running 
it on BMAX500 (F=0.34 \vs F=0.56 for AMAT).

\begin{table}
    \centering
    \begin{tabular}{|c|c|c|c|}
    \hline
    Metric	&	Precision	&	Recall	&	F-measure 	\\
    \hline
    MIL~\cite{tsogkas2012learning}	&	0.49	& 	0.55 	& 	0.52    \\
    \hline
    AMAT	&	0.52	& 	0.63 	& 	\textbf{0.57}    \\
    \hline
    Human   &	0.89	& 	0.66 	& 	0.77	\\
    \hline
    \end{tabular}
    \caption{Medial point detection on the BSDS500 val set.}
    \label{tab:experiments:medialpointdetection-BMAX500}
\end{table}

\subsection{Image Reconstruction}\label{sec:experiments:reconstruction}
\begin{figure*}[t]
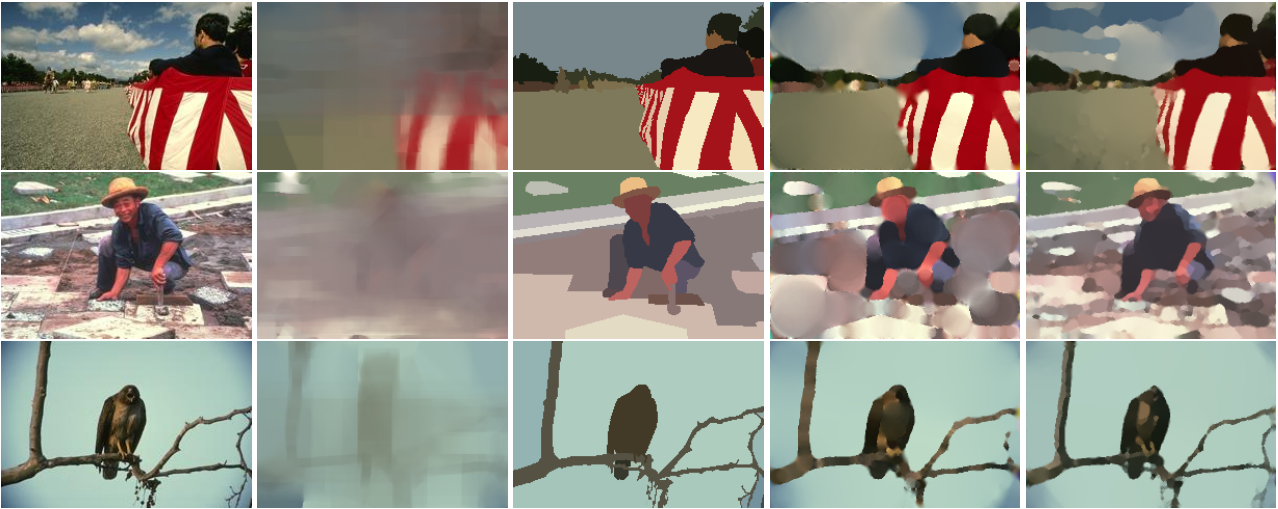

    \centering
    \def\imageWidth{0.19}

    \def\img_id{145086}
    \includegraphics[width=\imageWidth \textwidth]{\img_id_resized.jpg}
    \includegraphics[width=\imageWidth \textwidth]{\img_id_rec_mil.png}
    \includegraphics[width=\imageWidth \textwidth]{\img_id_rec_gtseg.png}
    \includegraphics[width=\imageWidth \textwidth]{\img_id_rec_gtskel.png}
    \includegraphics[width=\imageWidth \textwidth]{\img_id_rec_amat.png}

    \def\img_id{85048}
    \includegraphics[width=\imageWidth \textwidth]{\img_id_resized.jpg}
    \includegraphics[width=\imageWidth \textwidth]{\img_id_rec_mil.png}
    \includegraphics[width=\imageWidth \textwidth]{\img_id_rec_gtseg.png}
    \includegraphics[width=\imageWidth \textwidth]{\img_id_rec_gtskel.png}
    \includegraphics[width=\imageWidth \textwidth]{\img_id_rec_amat.png}

    \def\img_id{42049}
    \includegraphics[width=\imageWidth \textwidth]{\img_id_resized.jpg}
    \includegraphics[width=\imageWidth \textwidth]{\img_id_rec_mil.png}
    \includegraphics[width=\imageWidth \textwidth]{\img_id_rec_gtseg.png}
    \includegraphics[width=\imageWidth \textwidth]{\img_id_rec_gtskel.png}
    \includegraphics[width=\imageWidth \textwidth]{\img_id_rec_amat.png}

    \caption{
        \textbf{Image reconstruction}. From left to right: Input image, 
        MIL~\cite{tsogkas2012learning}, GT-seg, GT-skel, AMAT.
    }
    \label{fig:experiments:reconstruction}
\end{figure*}

We now assess the quality of reconstructions we obtain by inverting the computed AMAT
of images from the BSDS500 dataset.
We compare with a baseline reconstruction algorithm based on the MIL approach 
of~\cite{tsogkas2012learning} (after retraining MIL-color on BMAX500).
Their method uses features extracted in rectangular areas 
to produce a map of medial point strength at 13 scales and 8 orientations, for each pixel.
A single confidence value for each point is derived through a noisy-or operation,
which does away with scale and orientation information.
As a surrogate, in our experiments we associate each point with the scale/orientation combination
that has the highest score.

The scheme we use to create a crude reconstruction with their approach is the following:
We start by sorting medial point scores in decreasing order and we pick the highest-scoring point.
The rectangular region at the respective scale and orientation is then marked as covered,
and the process is repeated until the whole image has been reconstructed.
Similarly to our own method, point encodings are the mean RGB values within the rectangle, 
and local reconstructions are computed by averaging overlapping encodings.
We also compare with two more baselines: one obtained by considering ground-truth (GT) segments in BSDS500
and representing them by their mean RGB values (GT-seg); 
and a second, obtained through the GT skeletons and radii in BMAX500 (GT-skel).
For the latter, we use the reconstruction process described in~\refsec{sec:implementation:inverting}.

We consider three standard evaluation metrics for image similarity: MSE, PSNR, and SSIM.
Results are reported in~\reftab{tab:experiments:reconstruction} and visual 
examples are shown in~\reffig{fig:experiments:reconstruction}.
MIL uses rectangle filters at a finite set of scales and orientations that do 
not always match the scale and orientation of structures present in an image.
As a result, MIL reconstructions are very blurred.
GT-based reconstructions, on the other hand, have sharp edges but tend to have less
texture detail, because people tend to undersegment images, favoring \emph{perceptual} coherence 
over region appearance coherence.
Note that, for each image, we choose the GT annotation that produces the best SSIM score, 
to ensure we are always comparing against the best possible GT-based reconstruction.

\begin{table}
    \centering
    \begin{tabular}{|c|c|c|c|c|}
    \hline
    Metric	&	MSE		&	PSNR (dB)	&	SSIM	&	Compression 	\\
    \hline
    MIL~\cite{tsogkas2012learning}	&	0.0258	& 	16.6 	& 	0.53	&	$20\times$	\\
    \hline
    GT seg	&	0.0149	& 	18.87 	& 	0.64	&	$9\times$\\
    \hline
    GT skel &	0.0114	& 	20.19 	& 	0.67	&	$14\times$\\
    \hline
    AMAT	&	\textbf{0.0058}	&	\textbf{22.74}	&	\textbf{0.74}	&	$11\times$	\\
    \hline
    \end{tabular}
    \caption{Image reconstruction quality in BSDS500 val set.}
    \label{tab:experiments:reconstruction}
\end{table}
\section{Discussion}\label{sec:discussion}


We have defined the first complete medial axis transform for natural images.
Our approach bridges the gap between MAT methods for binary shapes and 
medial axis/local symmetry detection methods for real images.
We have demonstrated \sota\ performance in medial point detection and shown that 
we can produce a high-quality rendering of the input image using as few as $10\%$ of its pixels. 

That said, it is important to note that AMAT is not designed to be optimal for 
either of these tasks. 
Instead, it is designed to strike a balance between two conflicting objectives: 
i) capturing an image's salient structures (in the form of medial axes and 
their respective scale/appearance information); ii) providing an accurate
reconstruction of the original image from this abstracted representation. 
Therefore, performance should be assessed on both objectives jointly.

We also want to emphasize that AMAT is a purely bottom-up algorithm, completely 
unsupervised and train-free. 
We consider this an important advantage of our approach, as it means that it can 
generalize well and in a predictable way to new datasets, without the need for 
additional tuning. 
Despite the lack of training, we have shown that it performs surprisingly well, 
and can even be competitive with supervised methods fine-tuned to specific datasets.

In future work, our goal is to parameterize our method to accommodate the
relative roles of shape and appearance, and allow for flexible hierarchical 
grouping of medial branches to support segmentations of varying granularities.
Furthermore, although our current choice of $f/g$ favors simplicity and 
compactness  at the cost of texture, our framework can accommodate \emph{any} encoding/decoding functions.
Designing alternatives to better capture and reconstruct texture, or for specific discriminative tasks, is another exciting future direction.

\section*{Acknowledgements}\label{sec:acknowledgements}
This work was funded by NSERC. 
We thank Ioannis Gkioulekas for his valuable 
suggestions and feedback. 

{\small
\bibliographystyle{ieee}
\bibliography{iccv2017.bib}
}

\end{document}